\documentclass{article}

\PassOptionsToPackage{numbers, sort&compress}{natbib}

\usepackage[preprint]{neurips_2022}

\usepackage{graphicx}
\usepackage{amsmath}
\usepackage{amssymb}
\usepackage{booktabs}
\usepackage{caption}
\usepackage{subcaption}

\usepackage[pagebackref,breaklinks,colorlinks]{hyperref}

\usepackage[capitalize]{cleveref}
\crefname{section}{Sec.}{Secs.}
\Crefname{section}{Section}{Sections}
\Crefname{table}{Table}{Tables}
\crefname{table}{Tab.}{Tabs.}

\begin{document}

\title{NutritionVerse-Thin: An Optimized Strategy for Enabling Improved Rendering of 3D Thin Food Models}

\author{
Chi-en Amy Tai\textsuperscript{*}\textsuperscript{1}
\qquad Jason Li\textsuperscript{*}\textsuperscript{1}
\qquad Sriram Kumar\textsuperscript{*}\textsuperscript{1} \\
\qquad \textbf{Saeejith Nair\textsuperscript{1}}
\qquad \textbf{Yuhao Chen\textsuperscript{1}}
\qquad \textbf{Pengcheng Xi\textsuperscript{2}}
\qquad \textbf{Alexander Wong\textsuperscript{1}} \\
\textsuperscript{*} All authors contributed equally.\\
\textsuperscript{1} Vision and Image Processing Lab, University of Waterloo\\
\textsuperscript{2} National Research Council Canada \\
{\tt\small \{amy.tai, j2643li, ssriramk, smnair, yuhao.chen1, alexander.wong\}@uwaterloo.ca} \\
{\tt\small {pengcheng.xi}@nrc-cnrc.gc.ca}
}

\maketitle

\begin{abstract}
With the growth in capabilities of generative models, there has been growing interest in using photo-realistic renders of common 3D food items to improve downstream tasks such as food printing, nutrition prediction, or management of food wastage. Despite 3D modelling capabilities being more accessible than ever due to the success of NeRF based view-synthesis, such rendering methods still struggle to correctly capture thin food objects, often generating meshes with significant holes. In this study, we present an optimized strategy for enabling improved rendering of thin 3D food models, and demonstrate qualitative improvements in rendering quality. Our method generates the 3D model mesh via a proposed thin-object-optimized differentiable reconstruction method and tailors the strategy at both the data collection and training stages to better handle thin objects. While simple, we find that this technique can be employed for quick and highly consistent capturing of thin 3D objects.
\end{abstract}

\section{Introduction}
\begin{figure}[!ht]
    \begin{center}
    \subfloat[\centering instant-ngp]{{\includegraphics[width=.43\linewidth]{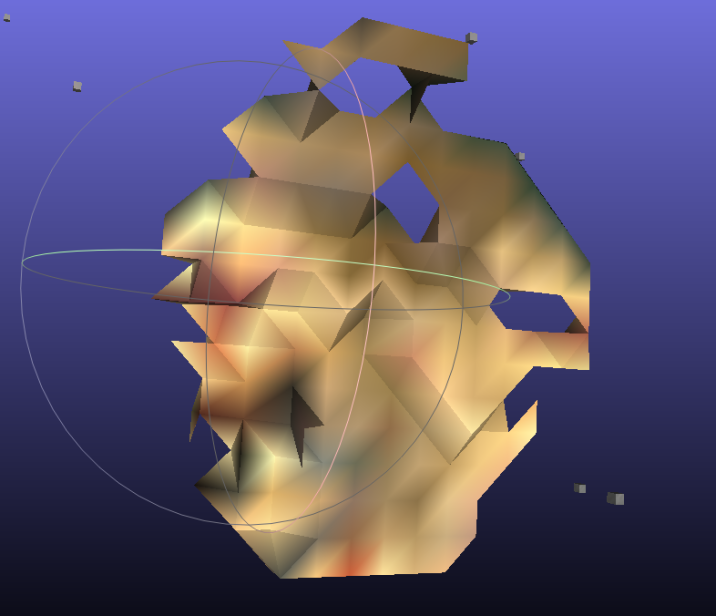} }}%
    \qquad
    \subfloat[\centering Nutriverse-Thin]{{\includegraphics[width=.41\linewidth]{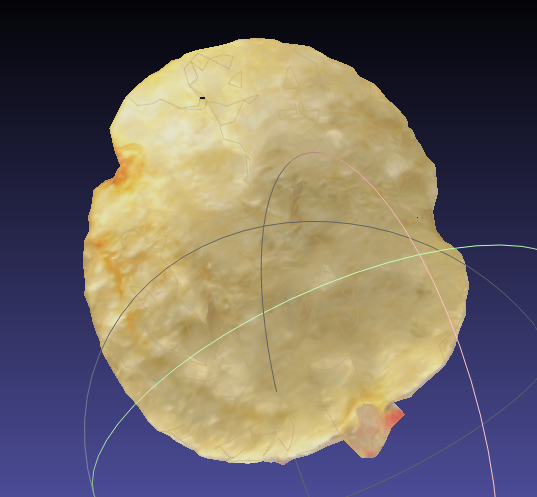} }}%
    \end{center}
    \caption{Comparison of a Lay's chip mesh representation rendered using instant-ngp (left) and Nutriverse-Thin (right).}
    \label{fig:difference-ngp-nutriverse}
\end{figure}

With the growth in capabilities of generative models, there has been growing interest in using photo-realistic renders of common 3D food items. Having 3D models of food is incredibly beneficial for food printing to help design foods with new textures that are potentially more nutritious \cite{3d-printing}. 3D food models can also assist with the generation of more data for volume estimation and nutrient prediction \cite{food-vol-estimation-3D, food-vol-est-phone}. Another use case for 3D food models is to improve management of food wastage as the internal structure of food can be modified to control user's food intake \cite{foodfab-food-wastage}. Despite 3D modelling capabilities being more accessible than ever due to the success of NeRF based view-synthesis, such rendering methods still struggle to correctly capture thin food objects, often generating meshes with significant holes, as seen in \cref{fig:difference-ngp-nutriverse}.

In this study, we present an optimized strategy for enabling improved rendering of thin 3D food models, and demonstrate qualitative improvements in rendering quality. Our method generates the 3D model mesh via a proposed thin-object-optimized differentiable reconstruction method \cite{munkberg2021nvdiffrec} and tailors the strategy at both the data collection and training stages to better handle thin objects. While simple, we find that this technique can be employed for quick and highly consistent capturing of 3D thin objects.

\section{Methodology}
\begin{figure*}[h]
    \begin{center}
        \includegraphics[width=\linewidth]{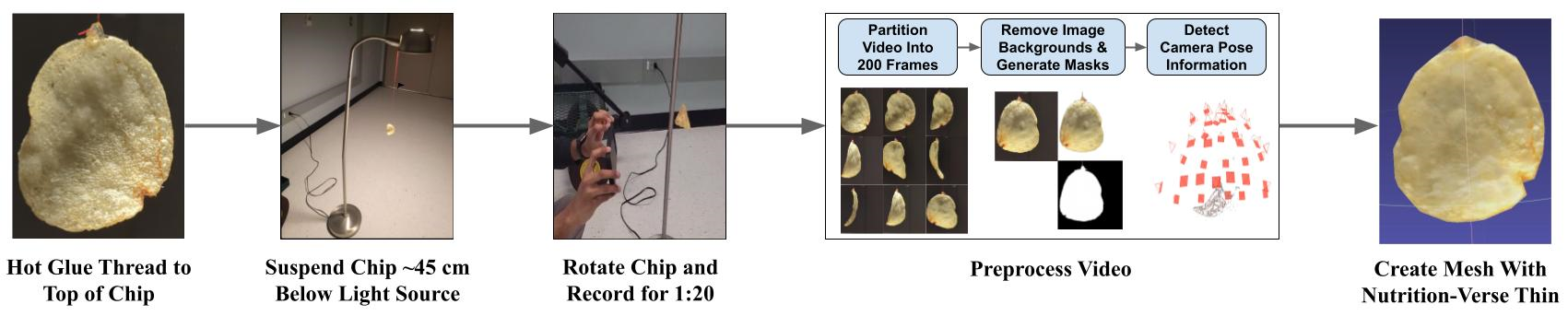}
    \end{center}
    \caption{Overview of the process map for an example thin food object using the proposed optimized strategy.}
    \label{fig:process-map}
\end{figure*}

The proposed optimized strategy for enabling improved rendering of 3D thin food models consists of optimizations in both the data collection and model training phases.  An overview of the entire process is shown in \cref{fig:process-map}. 

\subsection{Data Collection}
A significant problem that arises when trying to create a 3D model of a thin object is image capture from different viewpoints since the object is unable to stand vertically without external support. Providing the necessary support obscures much of the article and stands out in the background, preventing quality data collection of the object itself. To alleviate this problem and collect multi-view perspectives without obstruction, we first attach the object to a thread using hot glue and next suspend the thread from a platform. We then rotate the thread (and thus the object) to capture images of the object from all 360 degrees. 

Another issue is the existence of inconsistent lighting which results in irregularities in surface brightness. To minimize interference and ensure consistency, we suspend the object directly (45 cm) below a light source that shines perpendicular to the ground and dim all other light sources in the vicinity as seen in \cref{fig:process-map}. A Google Pixel 7 Pro phone is used to record a video of the rotating item; the item always remains centered in the frame. We position the camera parallel to the item at around 10-15 cm from the item as these close-up shots help capture maximum detail, thus optimizing render quality. We gradually move the phone down and subsequently up (while maintaining distance and centering) to capture the top and bottom of the article. The suggested video length is 1:20 as this duration is short enough to avoid lengthy preprocessing times while allowing for competent multi-view data collection.

Approximately 200 frames are then sampled from the video at equal intervals using OpenCV \cite{opencv_library}. The backgrounds of the sampled images are removed using Rembg and image masks are generated \cite{rembg}. The images are downscaled to a resolution of (512x512) to reduce strain on the GPU during the mesh extraction process. The preprocessed images are then inputted to COLMAP \cite{colmap} which produces intrinsic and extrinsic camera parameters of the accepted images \cite{mildenhall2019llff}. We find that performing the more comprehensive exhaustive feature matching (compared to sequential matching) produces better results without imposing a significant time delay due to our small image set. Finally, the pose information, images, and image masks are passed into a thin object-optimized differentiable reconstruction method \cite{munkberg2021nvdiffrec} for model training and mesh generation. 

\subsection{Model Training}
\begin{figure}[!ht]
    \begin{center}
    \subfloat[\centering Original Lay's Chip]{{\includegraphics[width=.4\linewidth]{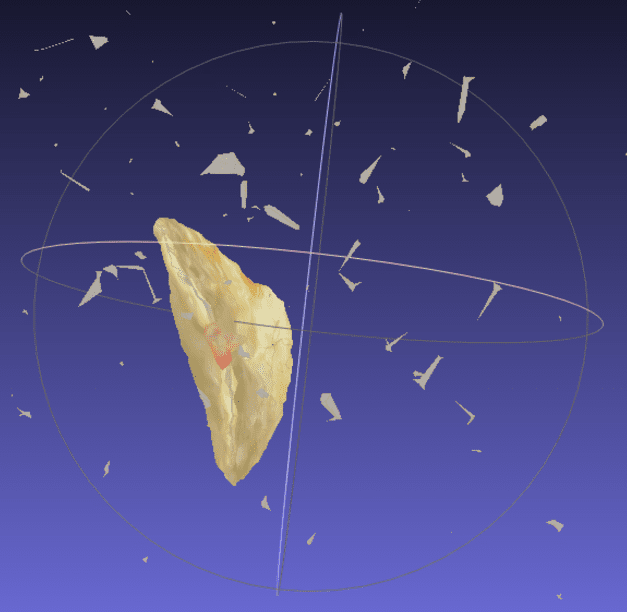} }}%
    \qquad
    \subfloat[\centering Optimized Lay's chip]{{\includegraphics[width=.39\linewidth]{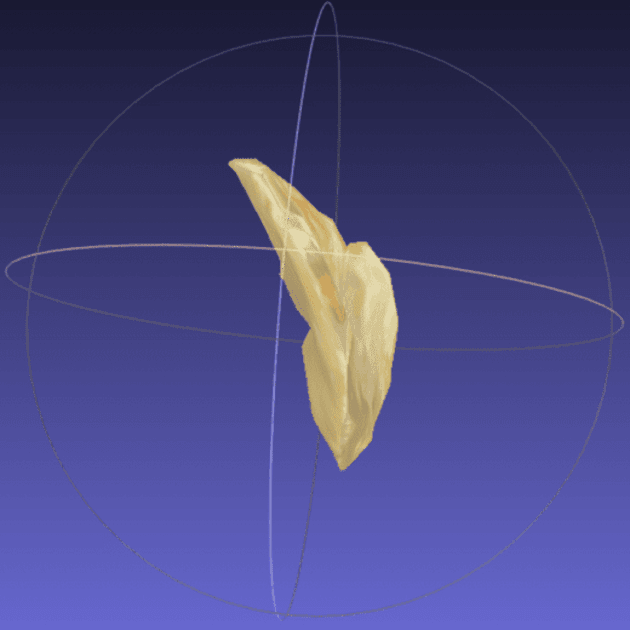} }}%
    \end{center}
    \begin{center}
    \subfloat[\centering Original Doritos chip\cite{munkberg2021nvdiffrec}]{{\includegraphics[width=.4\linewidth]{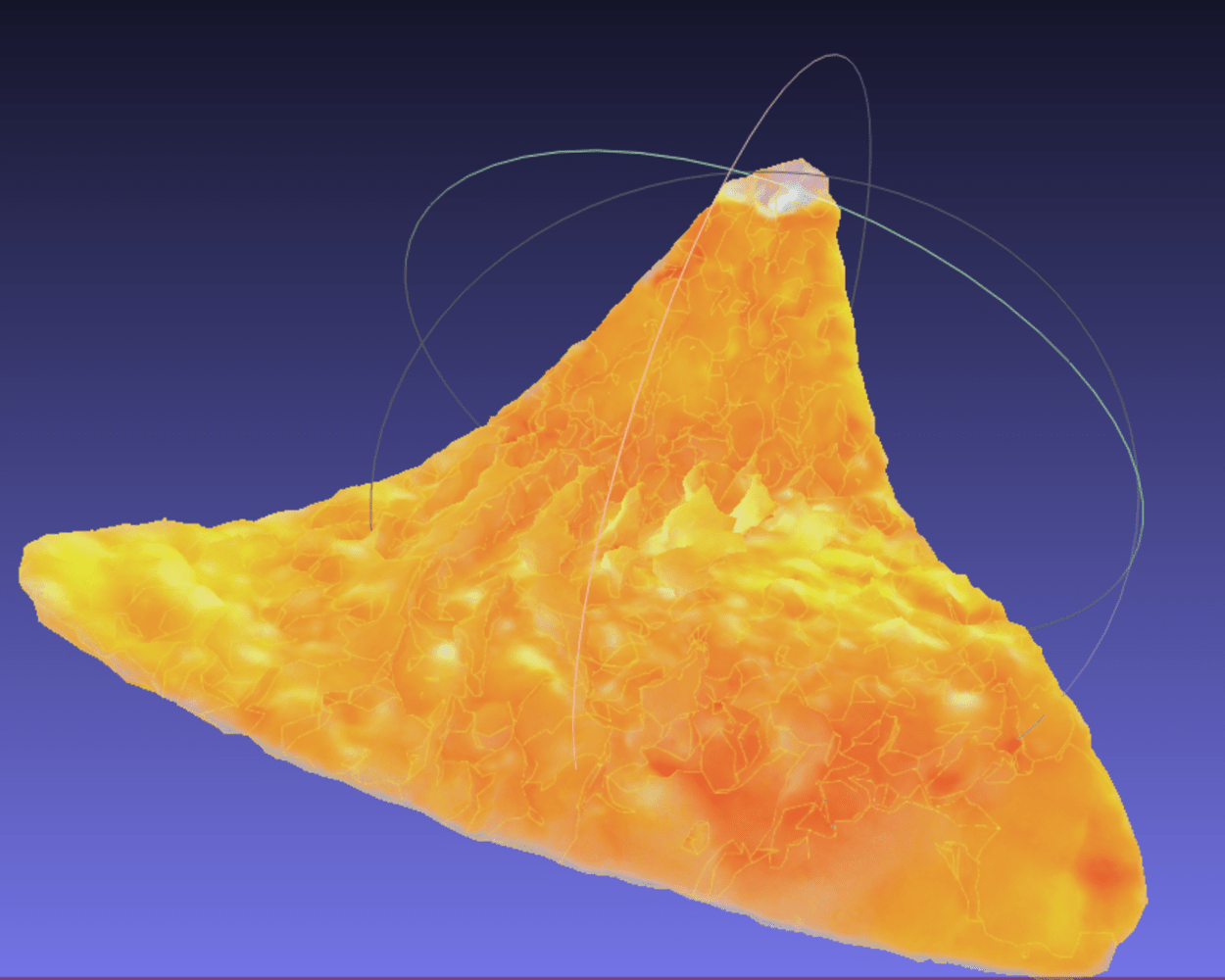} }}%
    \qquad
    \subfloat[\centering Optimized Doritos chip]{{\includegraphics[width=.4\linewidth]{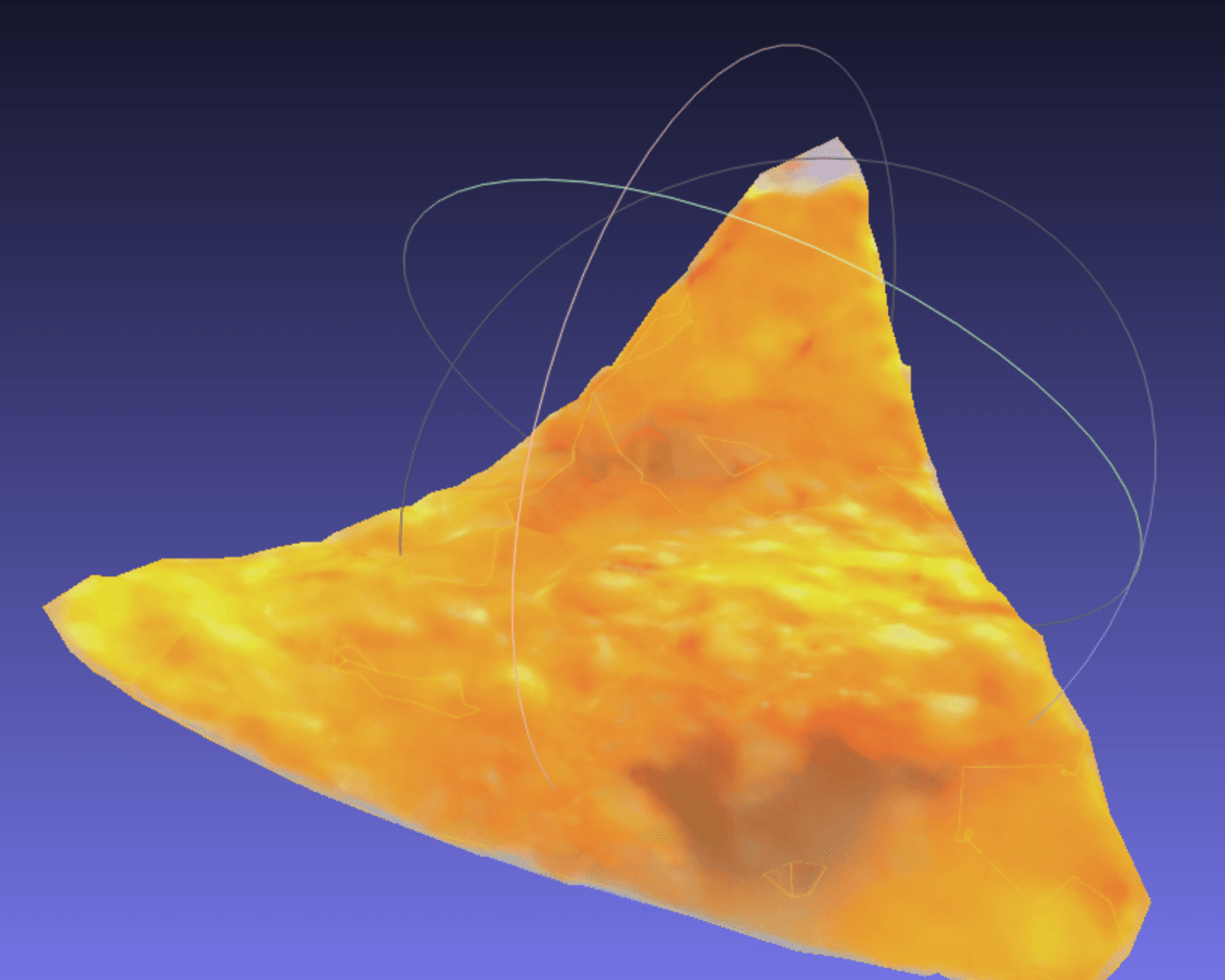} }}%
    \end{center}
    \caption{Comparisons of 3D model representations of a Lay's and Doritos chip generated using the original and thin-object-optimized differentiable reconstruction method.}
    \label{fig:nvdiffrec-before-after}
\end{figure}

We first investigated instant-ngp \cite{mueller2022instant} for rendering of thin objects but found that the outputted mesh had an irregular/rough surface and contained numerous holes as seen in \cref{fig:difference-ngp-nutriverse}. To address the poor rendering results of existing methods, we propose an extended training strategy for optimizing the differentiable reconstruction method proposed in \cite{munkberg2021nvdiffrec} to handle thin objects. More specifically, we conduct training with a reduced tetrahedral grid resolution, a large Laplacian value, and strong signed distance field regularization to enforce greater mesh smoothness. While these constraints do come at the expense of minor detail loss, they regularize the surface geometric shape of items and result in significant reductions in the jaggedness of the mesh. An additional benefit is the elimination of ghost effects as seen in \cref{fig:nvdiffrec-before-after}. Finally, we recommend a lower training resolution and number of iterations as these significantly lower computational costs  without incurring any noticeable differences in render quality.

\begin{figure}[!ht]
    \begin{center}
    \subfloat[\centering Sample Input Image]{{\includegraphics[width=.4\linewidth]{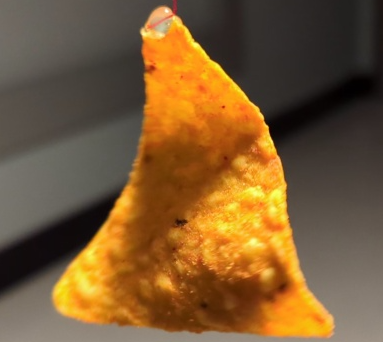} }}%
    \qquad
    \subfloat[\centering 3D Model Mesh]{{\includegraphics[width=.4\linewidth]{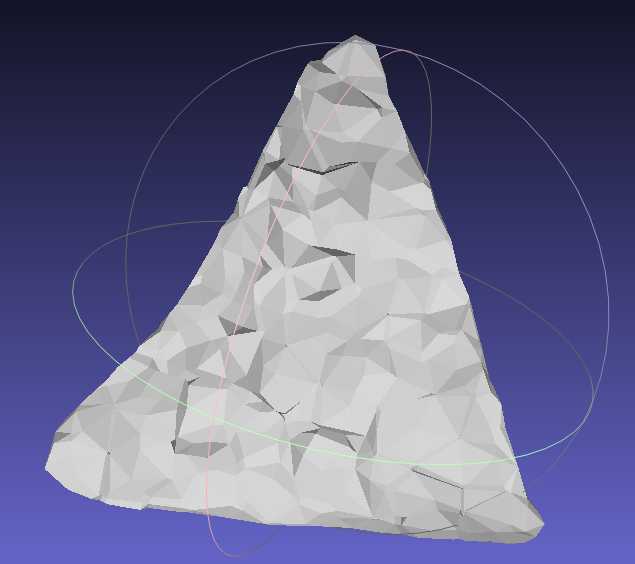} }}%
    \end{center}
    \caption{Example of a successful 3D model rendering from a sample input image.}
    \label{fig:nvdiffrec-doritos}
\end{figure}

\section{Conclusion}
In this paper, we present NutritionVerse-Thin, a technique that can be employed for quick and consistent capturing of 3D thin objects and show sample successful renderings of chips using this training strategy. However, this approach can be implemented for collecting other thin food models and hence, would be beneficial in advancing food printing, improving models for nutrient prediction, and reducing food wastage. 

\bibliographystyle{unsrtnat}
{
\small

\bibliography{main}
}

\end{document}